\documentclass[letterpaper]{article} 
\usepackage[submission]{aaai23}  
\usepackage{times}  
\usepackage{helvet}  
\usepackage{courier}  
\usepackage[hyphens]{url}  
\usepackage{graphicx} 
\usepackage{natbib}  
\usepackage{caption} 
\usepackage{algorithm}
\usepackage{algorithmic}

\title{ICPC: Instance-Conditioned Prompting with Contrastive Learning for Semantic Segmentation}

\author{Chaohui Yu,~~~~~~~~~~Qiang Zhou,~~~~~~~~~~Zhibin Wang,~~~~~~~~~~Fan Wang\\
Alibaba Group\\
{\tt \{huakun.ych,jianchong.zq,zhibin.waz,fan.w\}@alibaba-inc.com}}

\usepackage{bibentry}

\usepackage{amsmath}
\usepackage{booktabs}
\usepackage{multirow}
\usepackage{multicol}
\usepackage{amsfonts}
\usepackage{bbm}
\usepackage{mathtools}
\usepackage{subfigure}
\usepackage{bbding}
\usepackage{color}

\def\eg{{\itshape e.g.}}
\def\ie{{\itshape i.e.}}

\def\wrt{{\itshape w.r.t.}}

\begin{document}

\maketitle

\begin{abstract}
Modern supervised semantic segmentation methods are usually finetuned based on the supervised or self-supervised models pre-trained on ImageNet.
%
Recent work shows that transferring the knowledge from CLIP to semantic segmentation via prompt learning can achieve promising performance. 
The performance boost comes from the feature enhancement with multimodal alignment, \ie, the dot product between vision and text embeddings.
%
%
%
%
%
However, how to improve the multimodal alignment for better transfer performance in dense tasks remains underexplored.
In this work, we focus on improving the quality of vision-text alignment from two aspects of prompting design and loss function, and present an instance-conditioned prompting with contrastive learning (ICPC) framework.
First, compared with the static prompt designs, we reveal that dynamic prompting conditioned on image content can more efficiently utilize the text encoder for complex dense tasks.
Second, we propose an align-guided contrastive loss to refine the alignment of vision and text embeddings. 
We further propose lightweight multi-scale alignment for better performance. 
Extensive experiments on three large-scale datasets (ADE20K, COCO-Stuff10k, and ADE20K-Full) demonstrate that ICPC brings consistent improvements across diverse backbones. Taking ResNet-50 as an example, ICPC outperforms the state-of-the-art counterpart by 1.71\%, 1.05\%, and 1.41\% mIoU on the three datasets, respectively.



\end{abstract}

\section{Introduction}


Semantic segmentation is a fundamental task in computer vision and has witnessed great progress using deep learning in the past few years. It aims at segmenting things or stuff in an image as well as assigning category labels to them. For a long time, the ``pre-train and finetune'' training paradigm has dominated the deep learning-based computer vision tasks including semantic segmentation. 
Starting from FCN~\cite{long2015fully}, many semantic segmentation approaches have benefited from the supervised~\cite{he2016deep} or self-supervised~\cite{grill2020bootstrap,he2020momentum,wang2021densecl} pre-trained models on ImageNet. 

\begin{figure}[!t]
    \centering
    \includegraphics[width=0.96\linewidth]{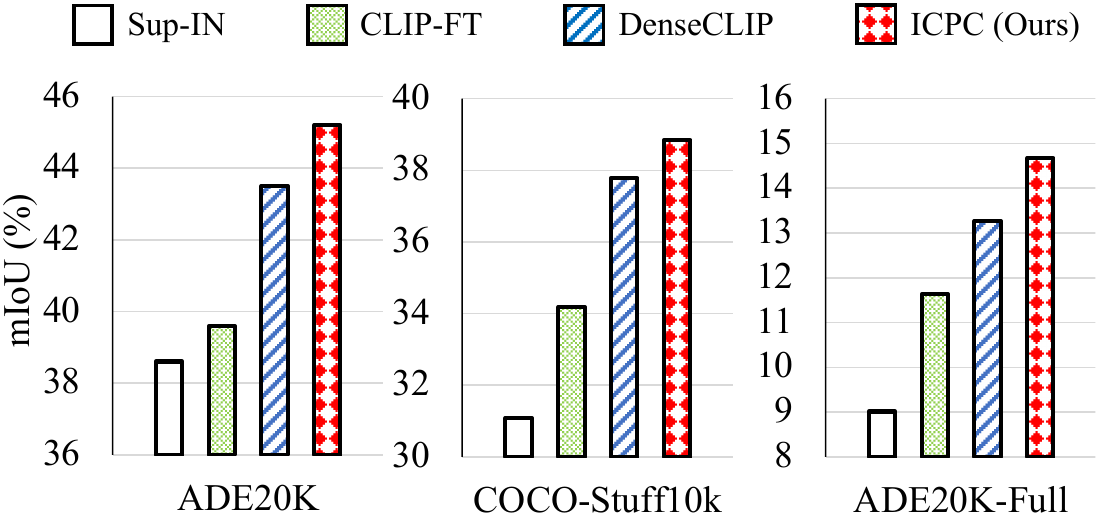}
    \vspace{-.1in}
    \caption{Comparison of finetuning-based (Sup-IN and CLIP-FT) and prompting-based (DenseCLIP~\cite{rao2022denseclip} and ICPC) methods. ``Sup-IN'' denotes the backbone is supervised pre-trained on ImageNet. ``CLIP-FT'' denotes using CLIP's image encoder as backbone.}
    \label{fig:intro_compare}
    \vspace{-.2in}
\end{figure}

Recently, vision-language models~\cite{radford2021learning,jia2021scalingalign} are bridging the gap between vision and text through large-scale pre-training. A representative method, CLIP~\cite{radford2021learning}, has been a promising alternative for visual pre-training these days.  
Inspired by the ``pre-train, prompt, and predict'' paradigm from NLP~\cite{liu2021promptnlp}, many prompt learning-based CLIP-driven methods have been proposed and can help to achieve state-of-the-art performance on classification tasks~\cite{zhou2021learning,zhou2022cocoop,gao2021clipadapter,zhang2021tipadapter}. 
%
%
Despite the progress in applying CLIP to classification tasks, how to transfer knowledge of vision-language models to supervised semantic segmentation is seldom explored. 
Very recently, DenseCLIP~\cite{rao2022denseclip} first proposes to apply CLIP to supervised dense tasks including semantic segmentation and demonstrates promising performance, indicating that multimodal alignment is also beneficial for dense tasks. 
%
%
%
%
Concretely, the performance boost of DenseCLIP comes from the feature enhancement with multimodal alignment, \ie, the dot product between vision and text embeddings, which is supervised by a per-pixel classification loss.
In other words, the quality of vision-text alignment is the key to leveraging the knowledge of CLIP in downstream semantic segmentation tasks.


However, neither the fixed ``{\tt a photo of a [CLASS]}'' nor the learnable form of prompt~\cite{zhou2021learning} used in DenseCLIP will be changed or updated once trained, which is too static to adaptively align the text and different image embeddings during inference, as depicted in Figure~\ref{fig:compare_tsne}(a).
Besides, only a per-pixel classification loss is used in DenseCLIP to supervise the alignment between vision and text embedding, which is considered weak to narrow the multimodal gaps since it limits to ``Vision $\rightarrow$ Text'' alignment, as depicted in Figure~\ref{fig:CL}.
We argue that such \textit{static} prompt designs and \textit{weak} vision-text alignment supervision will lead to coarse alignment between vision and text embeddings, thus limiting the transferability from CLIP~\cite{radford2021learning} to downstream semantic segmentation tasks.


In this work, we explore how to leverage the powerful knowledge of the CLIP model in semantic segmentation tasks more effectively. We focus on improving the alignment of text and visual embeddings and propose an \textbf{I}nstance-\textbf{C}onditioned \textbf{P}rompting with \textbf{C}ontrastive Learning framework, dubbed ICPC.

Firstly, we propose instance-conditioned prompting to encode the image-shared information as well as image-specific information for prompting.
Compared with the static prompt designs, we reveal that dynamic prompting conditioned on image content can more efficiently utilize the text encoder to describe different images adaptively.
Secondly, we introduce align-guided contrastive learning to refine the vision-text alignment, which attracts the positive pixel-text samples and dispels negative samples across alignments in a mini-batch by leveraging an easy-to-hard sampling strategy.
Compared with the weak alignment supervision that aligns vision to text via a segmentation loss, our align-guided contrastive loss performs alignment among multiple aligned image-text pairs, thus aligning ``Vision $\rightarrow$ Text'' and ``Text $\rightarrow$ Vision'' at the same time.
Finally, we propose lightweight multi-scale alignment to further align the text and visual object with different scales.

We evaluate ICPC on three large-scale semantic segmentation datasets with a broad range of categories: ADE20K~\cite{zhou2019semanticade20k} (150 classes), COCO-Stuff10k~\cite{caesar2018cocostuff} (171 classes), and ADE20K-Full~\cite{zhou2019semanticade20k} (847 classes). ICPC demonstrates superior performance on all three datasets. For instance, as shown in Figure~\ref{fig:intro_compare}, ICPC achieves the highest performance on all the three datasets based on ResNet-50+Semantic FPN, outperforming the finetuning-based methods and especially the prompting-based DenseCLIP by 1.71\%, 1.05\%, and 1.41\% mIoU, respectively.

Our main contributions are summarized as follows:
\begin{itemize}
\itemsep -0.1cm

\item We propose an instance-conditioned prompting with contrastive learning framework, called ICPC, to more effectively adapt the knowledge of the vision-language model (CLIP) to supervised semantic segmentation tasks.

\item To improve the alignment of vision and text embeddings, we propose two key improvements including instance-conditioned prompting and align-guided contrastive learning, and further introduce lightweight multi-scale alignment for better performance.

\item Extensive experiments on three large-scale datasets demonstrate that our ICPC framework outperforms previous finetuning-based methods and the state-of-the-art prompting-based DenseCLIP.
\end{itemize}

\section{Related Work}

\paragraph{Vision-Language Learning.} Vision-language learning plays an important role in the field of multimodal research, which cover a broad range of research topics, \eg, vision-language pre-training (VLP)~\cite{radford2021learning,jia2021scalingalign}, vision question answering (VQA)~\cite{antol2015vqa}, text-to-image generation~\cite{ramesh2022hierarchicaldalle2}, etc. We focus on reviewing the works on VLP and its applications on downstream tasks. VLP has achieved significant progress in the last few years, which jointly learns an image encoder and a text encoder and can be applied to many multimodal tasks. Among these VLP methods, a representative work is contrastive language-image pre-training, known as CLIP~\cite{radford2021learning}, which jointly trains the image and text encoders on 400 million image-text pairs collected from the web. 
Recently, CLIP shows that it can help to achieve state-of-the-art performance on few-shot or even zero-shot classification tasks~\cite{zhou2021learning,zhou2022cocoop,gao2021clipadapter,zhang2021tipadapter}. Inspired by prompt learning in NLP~\cite{liu2021promptnlp}, CoOp~\cite{zhou2021learning} and CoCoOp~\cite{zhou2022cocoop} propose to apply prompt learning to the adaptation of vision-language models in computer vision, which utilize learnable context vectors for better transfer performance. CLIP-adaptor~\cite{gao2021clipadapter} and TIP-adaptor~\cite{zhang2021tipadapter} propose to learn a lightweight network to improve the transferability of CLIP on downstream tasks. 
Different from these methods that focus on classification tasks, very recently, DenseCLIP~\cite{rao2022denseclip} proposes to apply CLIP to dense prediction tasks via language-guided finetuning and achieves promising results. 
This paper takes inspiration from these methods and focuses on how to leverage the knowledge of CLIP in supervised semantic segmentation tasks more effectively.

\paragraph{Contrastive Learning.} Recent years have witnessed great progress of self-supervised learning (SSL), in which, 
contrastive learning has been the most compelling method and dominates unsupervised representation learning. The contrastive methods learn representations in a discriminative manner by attracting similar (positive) instances and dispelling dissimilar (negative) instances. These methods usually generate two or multiple views of each image via strong augmentations to create positive instance pairs. 
For instance, InfoNEC~\cite{oord2018infonce} proposes a probabilistic contrastive loss to capture information that is maximally useful to predict future samples. 
SimCLR~\cite{chen2020simple} proposes to improve the performance of SSL by maximizing the mutual information between two augmented views of an image. 
MoCo~\cite{he2020momentum} utilizes a momentum encoder to maintain consistent representations of negative pairs drawn from a memory bank, 
which facilitates contrastive unsupervised learning. 
In addition to instance-level contrastive methods, DenseCL~\cite{wang2021densecl} proposes to perform pixel-level contrastive learning for unsupervised representation learning.

In this paper, we propose an align-guided contrastive objective to better align the vision and text in multimodal space.

\begin{figure*}[!t]
    \centering
    \includegraphics[width=0.9\linewidth]{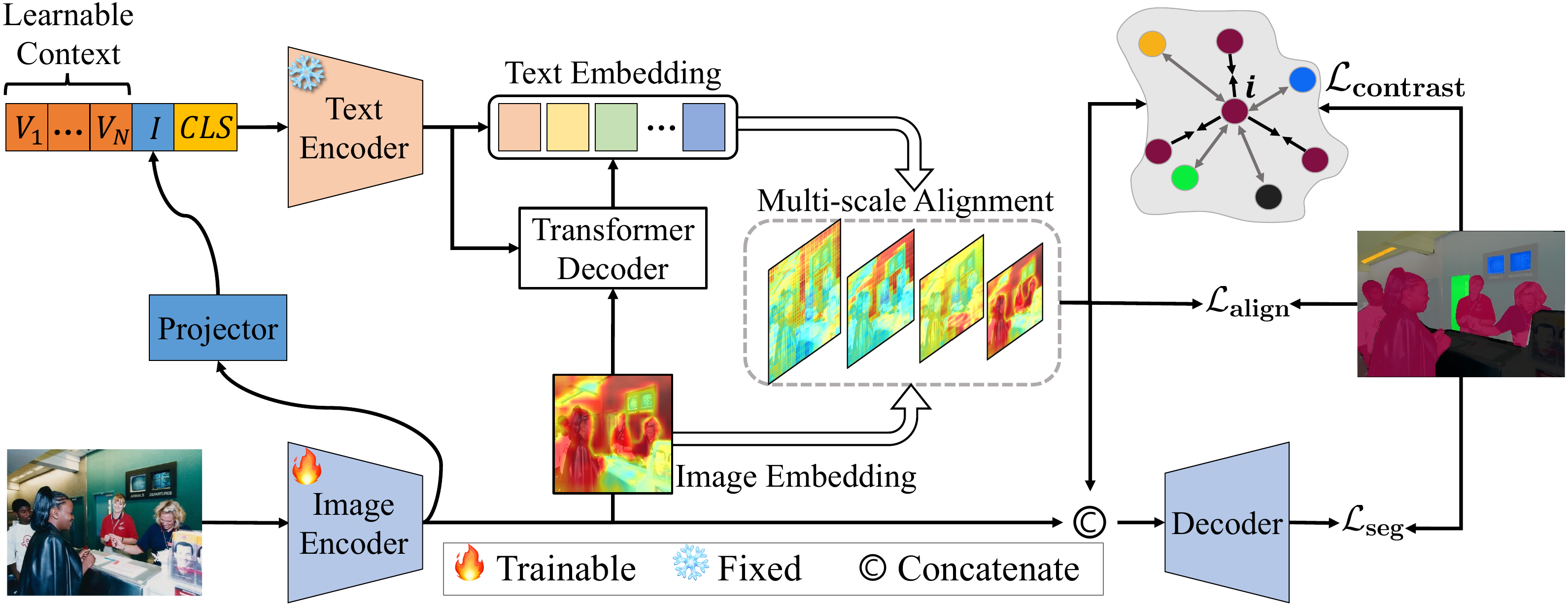}
    \vspace{-.1in}
    \caption{The overview of the proposed ICPC framework.}
    \label{fig:arch}
    \vspace{-.2in}
\end{figure*}

\paragraph{Semantic Segmentation.} Semantic segmentation~\cite{long2015fully,xiao2018unified,kirillov2019panoptic} aims to segment different target region (\eg, stuff or thing) in an image. FCN~\cite{long2015fully} first proposes a fully convolutional network that can be trained end-to-end, which greatly influences the later research on semantic segmentation. 
Modern semantic segmentation methods try to aggregate long-range information to explicitly encode the context. For instance, PSPNet~\cite{zhao2017pyramid} and Deeplab~\cite{chen2018encoder} use convolutional or pooling operators with different kernel sizes to gather multi-scale features. DANet~\cite{fu2019danet} and CCNet~\cite{huang2019ccnet} leverage different non-local attention~\cite{wang2018nonlocal} to model the context.
Recently, some methods~\cite{zheng2021rethinking,strudel2021segmenter} use Vision Transformer (ViT)~\cite{dosovitskiy2020imagevit} to model long-range context and achieved promising performance.

We place our work in the supervised semantic segmentation setting and build our framework upon CoOp~\cite{zhou2021learning} and DenseCLIP~\cite{rao2022denseclip}.

\section{Methodology}

 
We first revisit the CLIP method, which is the base model used in our method. Then, we present our ICPC framework as well as the rationale behind each component.

\subsection{Preliminaries: Review of CLIP}
Contrastive Language-Image Pre-training, known as CLIP~\cite{radford2021learning}, demonstrates a promising potential to explicitly leverage human language for solving computer vision tasks efficiently. CLIP mainly consists of two encoders: text encoder and image encoder. The image encoder can be ResNet~\cite{he2016deep} or ViT~\cite{dosovitskiy2020imagevit}, and the text encoder can be a Transformer~\cite{vaswani2017attention}. During pre-training, CLIP trains the image and text encoders on 400 million web-scale image-text pairs via a contrastive loss. To be specific, given a mini-batch of image-text pairs, CLIP aims to maximize the cosine similarities between the matched pairs while minimizing the similarities between mismatched pairs. After pre-training, the knowledge of CLIP can be transferred to various downstream recognition tasks. Specifically, let $x$ be the input image of image encoder $f_I$ and $t_y$ be the input prompt of text encoder $f_T$, such as ``{\tt a photo of a [CLASS]}'' where ``{\tt [CLASS]}'' is replaced with the $y$-th class name. The predicted classification probability is calculated as:
\begin{equation}
p(y|x) = \frac{\text{exp}(\text{cos}(f_I(x), f_T(t_y)) / \tau)} {\sum^{K}_{i=1} \text{exp}(\text{cos}(f_I(x), f_T(t_i))/\tau)},
\label{eq:clip_infer}
\end{equation}
where $\text{cos}(\cdot,\cdot)$ denotes the cosine similarity, $\tau$ is a learned temperature parameter, and $K$ is the total categories of a given dataset.

\subsection{Instance-Conditioned Prompting}


DenseCLIP~\cite{rao2022denseclip} adapts CoOp~\cite{zhou2021learning} as the base prompting, which shows that the learnable context also outperforms the fixed form of context for dense prediction tasks.
However, as shown in Figure~\ref{fig:compare_tsne}, we visualize the output image and text embeddings of the two encoders via T-SNE on the ADE20K validation set. The static prompt used in DenseCLIP~\cite{rao2022denseclip} can not align the image and text embeddings well.

Inspired by CoOp~\cite{zhou2021learning}, we also introduce learnable context to the prompt in our framework. In addition, to enable prompt to dynamically adapt to different input images, thus better aligning the vision and text modalities, we propose instance-conditioned prompting. Concretely, based on the $N$ learnable context vectors, denoted as $[V_1, V_2, ..., V_N]$, we further introduce a dynamic vector conditioned on image content, denoted as $I$, to the context. The final prompt to the text encoder can be formulated as:
\begin{equation}
P_k = [V_1, V_2, ..., V_N, I, \text{CLS}_k], ~~ 1 \leq k \leq K,
\label{eq:prompt}
\end{equation}
where $K$ denotes the total categories, $[V_1, V_2, ..., V_N] \in \mathbb{R}^{1 \times C}$ and $I \in \mathbb{R}^{1 \times C}$ are the learnable contexts and instance-conditioned context respectively, in which $N$ (8 by default) is the number of context vectors and $C$ (512 by default) is the dimension of each vector. $\text{CLS}_k$ is the embedding for the name of the $k$-th class via BPE~\cite{sennrich2015bpe}.
To attain the instance-conditioned context $I$, we propose to learn a lightweight network (the projector in Figure~\ref{fig:arch}) to generate a corresponding vector for each image. Let $f_P$ be the projector, then $I$ is computed as:
\begin{equation}
I = f_P ( f_I ( x )_{\text{global}} ),
\label{eq:instance_vector}
\end{equation}
where $f_I ( x )_{\text{global}} \in \mathbb{R}^{1 \times D}$ indicates that we only use the global feature of image encoder $f_I$ from CLIP. As for the projector, we use a simple two-layer design that consists of two $\text{linear}$ layers to project the global feature from $\mathbb{R}^{1 \times D}$ to $\mathbb{R}^{1 \times C}$.

\def\widvis{0.4}
\begin{figure}[t!]
\centering
\subfigure[DenseCLIP]{
\includegraphics[width=\widvis\linewidth]{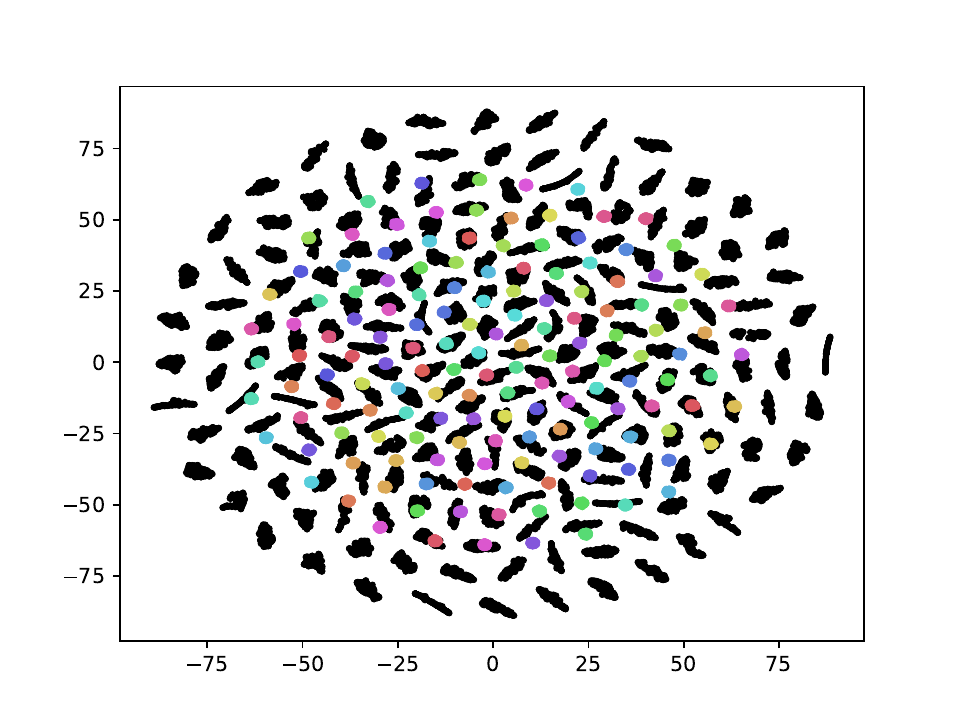}}
\quad
\subfigure[ICPC (Ours)]{
\includegraphics[width=\widvis\linewidth]{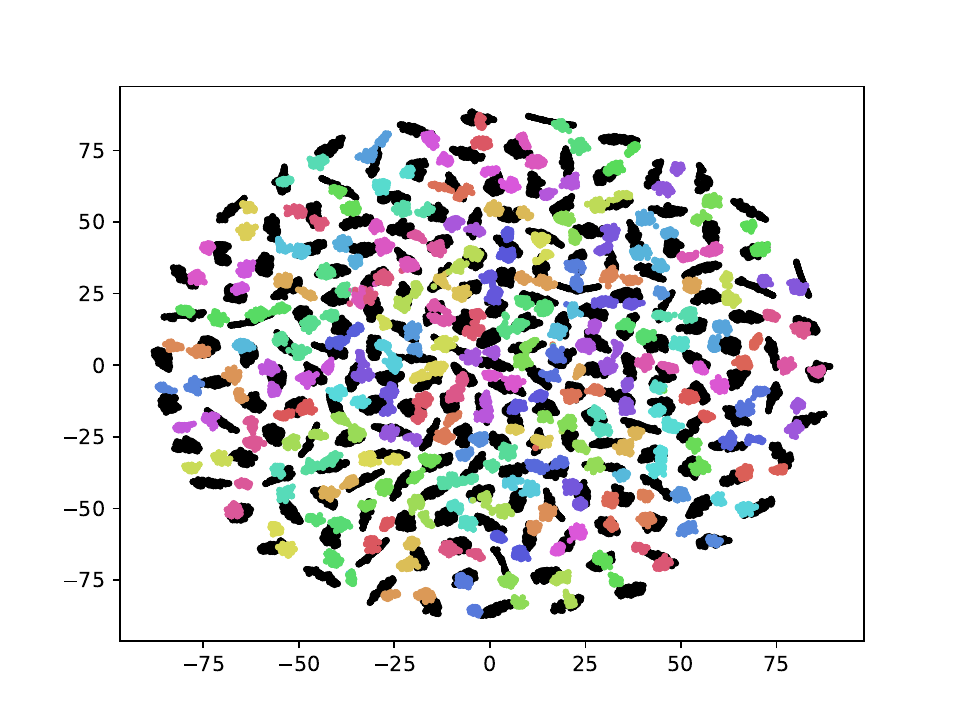}}
\vspace{-.1in}
\caption{T-SNE comparison of image embeddings (black) and text embeddings (each color denotes one class) on the validation set of ADE20K.}
\label{fig:compare_tsne}
\vspace{-.2in}
\end{figure}

Given the text embedding $\mathcal{T} = f_T ( P_k )$ and the image embedding $\mathcal{I} = f_I ( x )$, we follow the interaction design between these two modalities of~\cite{rao2022denseclip} to further refine the text embeddings. Specifically, we first perform cross-attention with a Transformer decoder~\cite{vaswani2017attention} by taking as input the text embedding as query and image embedding as key and value. Then, we update the text embedding as:
\begin{equation}
\mathcal{T} = \mathcal{T} + \lambda \cdot {\tt cross\_attn}(\mathcal{T}, \mathcal{I}),
\label{eq:textemb_update}
\end{equation}
where $\lambda$ is learnable trade-off parameter ($1e^{-4}$ by default).

The learnable context $[V_1, V_2, ..., V_N]$ shared among all images is considered to encode the general information of the entire dataset, while the instance-conditioned vector $I$ is considered to encode the image-specific information that can dynamically depict the image more accurately.
``\textit{A picture is worth a thousand words}'', the $I$ contains more information about the image, \eg, appearance, shape, etc.
As shown in Figure~\ref{fig:compare_tsne}(b), 
our instance-conditioned prompting is more refined and can better align the vision and text embeddings, which is beneficial for downstream dense tasks.

\subsection{Align-Guided Contrastive Learning}

\begin{figure}[!t]
    \centering
    \includegraphics[width=0.87\linewidth]{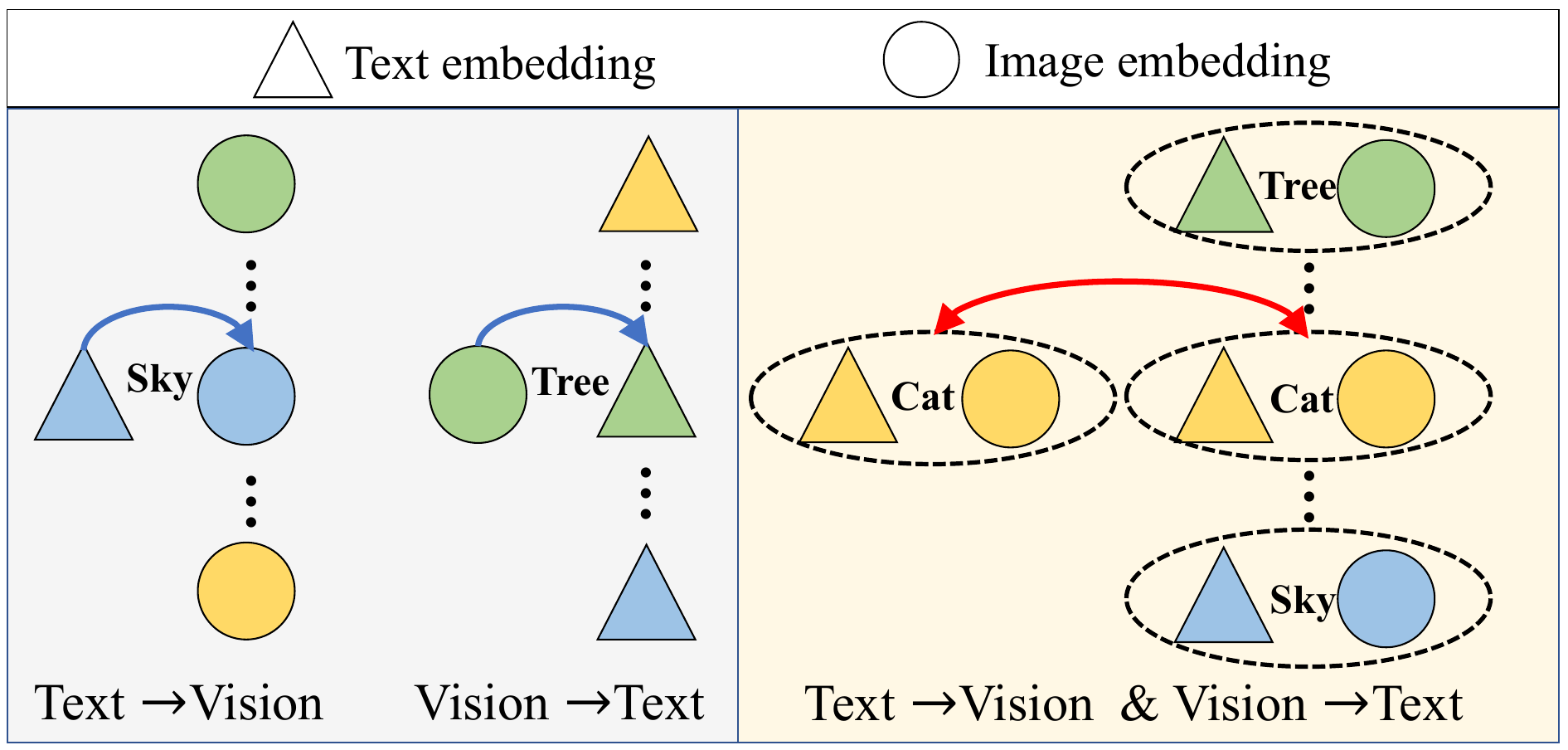}
    \caption{Illustration of the differences in vision-text alignment between existing methods (left) and our ICPC (right).}
    \label{fig:CL}
    \vspace{-.2in}
\end{figure}

Given the image embedding $\mathcal{I} \in \mathbb{R}^{H \times W \times C}$ and text embedding $\mathcal{T} \in \mathbb{R}^{K \times C}$, we then perform vision-text alignment by:
\begin{equation}
\mathcal{A} = \mathcal{I} \cdot \mathcal{T}^T, ~~ \mathcal{A} \in \mathbb{R}^{H \times W \times K}
\label{eq:dense_pred}
\end{equation}
where $\mathcal{A}$ indicates that each visual pixel is assigned a score across all $K$ categories. 
Based on the multimodal embeddings, DenseCLIP~\cite{rao2022denseclip} proposes to align each visual pixel embedding to $K$ class text embeddings via a per-pixel classification loss ($\mathcal{L}_{\text{align}}$ in Figure~\ref{fig:arch}), which is considered to perform ``Vision $\rightarrow$ Text'' alignment as depicted in Figure~\ref{fig:CL}.
%
However, the ``Text $\rightarrow$ Vision'' alignment is thus ignored.

Inspired by the recent advance in contrastive learning, we propose an align-guided contrastive objective to refine the vision-text alignment. 
Briefly speaking, based on multiple multimodal embeddings, we further perform alignment among these multiple aligned image-text pairs in a mini-batch, as depicted in the right part of Figure~\ref{fig:CL}. That is, in addition to ``Vision $\rightarrow$ Text'' alignment, our method implicitly takes the ``Text $\rightarrow$ Vision'' alignment into consideration.

Specifically, for a multimodal alignment point $p\in \mathbb{R}^{K}$ of $\mathcal{A}$ and its corresponding ground-truth class $k$, we collect all the points with the same class $k$ from the current mini-batch to compose positive samples $\mathbf{P}$, and collect the points with other classes to compose negative samples $\mathbf{N}$. Our align-guided contrastive loss is formulated as an InfoNCE loss~\cite{oord2018infonce} on these multimodal alignment points. For simplicity, we only show the contrastive objective on one point:
\begin{equation}
\begin{aligned}
& \mathcal{L}_{\text{contrast}} \\
& = \frac{1}{|\mathbf{P}|} \sum_{q_{+} \in \mathbf{P}}-\text{log} \frac{\text{exp}(p \cdot q_{+}/\tau)} {\text{exp}(p \cdot q_{+}/\tau) + \sum\limits_{q_{-} \in \mathbf{N}} \text{exp}(p \cdot q_{-}/\tau)},
\end{aligned}
\label{eq:contrast_loss}
\end{equation}
%
where $\tau$ is a temperature term (0.1 by default). For training efficiency and better performance, we only sample 5 points ($|\mathbf{P}|$ = 5) for each contained class from all the alignments $\mathcal{A}$ of a mini-batch.

In addition, to better use the alignment result to guide the contrastive learning process, we propose an easy-to-hard sampling strategy. 
Given the alignment result $\mathcal{A}$ and ground-truth labels, we first select those correctly classified points as easy samples and misclassified points as hard samples. Then we take inspiration from hard mining~\cite{kalantidis2020hard} and pay more attention to hard samples along the training progress, \ie, we choose more easy samples during the early phase for stable convergence, and we choose more and more hard samples as training. We use a simple linear schedule to perform easy-to-hard sampling:
\begin{equation}
\left\{
\begin{aligned}
|\mathbf{P}|_{hard} & = \frac{t}{T} \cdot |\mathbf{P}|, \\
|\mathbf{P}|_{easy} & = |\mathbf{P}| - |\mathbf{P}|_{hard},
\end{aligned}
\right.
\label{eq:easy2hard}
\end{equation}
where $t$ and $T$ denote the current and total training iterations, respectively.
The hard samples of $\mathcal{A}$ somehow indicate poor alignment between pixel and text embeddings and can be better refined with our align-guided contrastive loss, which is helpful for semantic segmentation tasks.

\subsection{Multi-Scale Vision-Text Alignment}


Multi-scale technique~\cite{lin2017featurefpn} has greatly promoted the advance of computer vision tasks. Previous work~\cite{rao2022denseclip} only uses features of the last stage to perform vision-text matching. However, we believe that aligning the embeddings of the textual class and the corresponding visual object with various scales can benefit the downstream dense tasks. We thus propose a simple lightweight design to perform multi-scale alignment:
\begin{equation}
\left\{
\begin{aligned}
\mathcal{A}_{\frac{1}{32}} & = \mathcal{I} \cdot \mathcal{T}^T, \\
\mathcal{A}_{\frac{1}{16}} & = \text{Up}_{\text{2X}}(\mathcal{I}) \cdot \mathcal{T}^T, \\
\mathcal{A}_{\frac{1}{8}}  & = \text{Up}_{\text{4X}}(\mathcal{A}_{\frac{1}{32}}) + \text{Up}_{\text{2X}}(\mathcal{A}_{\frac{1}{16}}), \\
\mathcal{A}_{\frac{1}{4}}  & = \text{Up}_{\text{8X}}(\mathcal{A}_{\frac{1}{32}}) + \text{Up}_{\text{4X}}(\mathcal{A}_{\frac{1}{16}}) + \text{Up}_{\text{2X}}(\mathcal{A}_{\frac{1}{8}}), \\
\end{aligned}
\right.
\label{eq:ms_align}
\end{equation}
where we use transposed convolution as the upsampling operator $\text{Up}(\cdot)$. These multi-scale alignments will be concatenated with the image embeddings as input to the decoder.

\subsection{Overall Learning Objective}
The overall learning objective consists of three parts: the original segmentation loss $\mathcal{L}_{\text{seg}}$, the vision-text alignment loss $\mathcal{L}_{\text{align}}$, and the align-guided contrastive loss $\mathcal{L}_{\text{contrast}}$:
\begin{equation}
\mathcal{L} = \underbrace{\mathcal{L}_{\text{seg}} + \mathcal{L}_{\text{align}}}_{\mathcal{L}_{\text{DenseCLIP}}} + \gamma \cdot \mathcal{L}_{\text{contrast}},
\label{eq:overallloss}
\end{equation}
where $\gamma$ is a balancing coefficient (0.5 by default). $\mathcal{L}_{\text{seg}}$ and $\mathcal{L}_{\text{align}}$ are inherited from DenseCLIP~\cite{rao2022denseclip} and are both per-pixel classification losses.

\section{Experiments}
\subsection{Datasets}


\noindent
\textbf{ADE20K}~\cite{zhou2019semanticade20k} is a large-scale semantic segmentation dataset, which contains a range of 150 categories. ADE20K consists of 20,210 training images and 2,000 validation images. Recently, most of the proposed literature for semantic segmentation are evaluated on ADE20K dataset. 

\noindent
\textbf{COCO-Stuff10k}~\cite{caesar2018cocostuff} is another large-scale semantic segmentation dataset, which consists of 10,000 images collected from COCO~\cite{lin2014microsoftcoco}. COCO-Stuff10k has 9,000 training images and 1,000 test images, and it provides rich annotations for 171 categories (80 for object, 91 for stuff).
    
\noindent
\textbf{ADE20K-Full}~\cite{zhou2019semanticade20k} is a more challenging dataset that is composed of 27,574 images (25,574 for training and 2,000 for testing) spanning 365 different scenes. Images are fully annotated with objects and many of the images also contain object parts.
Following ~\cite{cheng2021maskformer}, we keep 874 classes that are present in both training and validation sets.


\def\widvis{0.3cm}
\def\widviss{0.2cm}
\begin{table}[!t]
\centering
\resizebox{1.0\linewidth}{!}{
\begin{tabular}{clcll}
\toprule
Backbone                    & Method                   & Pre-train              & \hspace{\widvis}mIoU (\%)   & \hspace{\widviss}GFLOPs \\
\midrule
\multirow{11}{*}{ResNet-50} & FCN~\cite{long2015fully}                      & IN1K  & \hspace{\widvis}36.1 / 38.1      & \hspace{\widviss}793.3  \\
                            & EncNet~\cite{zhang2018context}                   & IN1K  & \hspace{\widvis}40.1 / 41.7      & \hspace{\widviss}565.6  \\
                            & PSPNet~\cite{zhao2017pyramid}                   & IN1K  & \hspace{\widvis}41.1 / 41.9      & \hspace{\widviss}716.2  \\
                            & CCNet~\cite{huang2019ccnet}                    & IN1K  & \hspace{\widvis}42.1 / 43.1      & \hspace{\widviss}804.0  \\
                            & DeeplabV3+~\cite{chen2018encoder}               & IN1K  & \hspace{\widvis}42.7 / 43.8      & \hspace{\widviss}711.5  \\
                            & UperNet~\cite{xiao2018unified}                  & IN1K  & \hspace{\widvis}42.1 / 42.8      & \hspace{\widviss}953.2  \\
                            & DNL~\cite{yin2020disentangled}                      & IN1K  & \hspace{\widvis}41.9 / 43.0      & \hspace{\widviss}939.3  \\
                            & Semantic FPN~\cite{kirillov2019panoptic}             & IN1K  & \hspace{\widvis}38.6 / 40.6      & \hspace{\widviss}227.1  \\ \cmidrule(lr){2-5}
                            & CLIP + Semantic FPN                               & CLIP      & \hspace{\widvis}39.6 / 41.6      & \hspace{\widviss}248.8  \\
                            & DenseCLIP~\cite{rao2022denseclip}                 & CLIP          & \hspace{\widvis}43.5 / 44.7      & \hspace{\widviss}269.7  \\
                            & ICPC (Ours)                                       & CLIP & \hspace{\widvis}\textbf{45.2} / \textbf{46.6}         &    \hspace{\widviss}364.0 (284.2$^{*}$)   \\
\midrule
\multirow{12}{*}{ResNet-101} & FCN~\cite{long2015fully}                      & IN1K  & \hspace{\widvis}39.9 / 41.4      & \hspace{\widviss}1104.4  \\
                             & EncNet~\cite{zhang2018context}                   & IN1K  & \hspace{\widvis}42.6 / 44.7      & \hspace{\widviss}876.8  \\
                             & PSPNet~\cite{zhao2017pyramid}                   & IN1K  & \hspace{\widvis}43.6 / 44.4      & \hspace{\widviss}1027.4  \\
                             & CCNet~\cite{huang2019ccnet}                    & IN1K  & \hspace{\widvis}44.0 / 45.2      & \hspace{\widviss}1115.2  \\
                             & DeeplabV3+~\cite{chen2018encoder}               & IN1K  & \hspace{\widvis}44.6 / 46.1      & \hspace{\widviss}1022.7  \\
                             & UperNet~\cite{xiao2018unified}                  & IN1K  & \hspace{\widvis}43.8 / 44.8      & \hspace{\widviss}1031.0  \\
                             & OCRNet~\cite{yuan2020object}                 & IN1K     & \hspace{\widvis}45.3 / \hspace{0.2cm}--         & \hspace{\widviss}923.9   \\
                             & DNL~\cite{yin2020disentangled}                      & IN1K  & \hspace{\widvis}44.3 / 45.8      & \hspace{\widviss}1250.5  \\
                             & Semantic FPN~\cite{kirillov2019panoptic}             & IN1K  & \hspace{\widvis}40.4 / 42.3      & \hspace{\widviss}304.9  \\ \cmidrule(lr){2-5}
                             & CLIP + Semantic FPN                                  & CLIP      & \hspace{\widvis}42.7 / 44.3      & \hspace{\widviss}326.6  \\
                             & DenseCLIP~\cite{rao2022denseclip}     & CLIP      & \hspace{\widvis}45.1 / 46.5      & \hspace{\widviss}346.3  \\
                             & ICPC (Ours)                           & CLIP      &   \hspace{\widvis}\textbf{47.1} / \textbf{48.5}         &  \hspace{\widviss}436.9 (357.1$^{*}$)   \\
\midrule
\multirow{6}{*}{ViT-B}       & SETR-MLA-DeiT~\cite{zheng2021rethinking}            & IN1K  & \hspace{\widvis}46.2 / 47.7      & \hspace{0.4cm}--  \\
                             & Semantic FPN~\cite{kirillov2019panoptic}             & IN1K  & \hspace{\widvis}48.3 / 50.9      & \hspace{\widviss}863.5  \\
                             & Semantic FPN~\cite{kirillov2019panoptic}             & IN21K  & \hspace{\widvis}49.1 / 50.4      & \hspace{\widviss}863.5 \\ \cmidrule(lr){2-5}
                             & CLIP + Semantic FPN                                  & CLIP      & \hspace{\widvis}49.4 / 50.3      & \hspace{\widviss}863.5  \\
                             & DenseCLIP~\cite{rao2022denseclip}     & CLIP      & \hspace{\widvis}50.6 / 51.3      & \hspace{\widviss}869.2  \\
                             & ICPC (Ours)                           & CLIP      &  \hspace{\widvis}\textbf{51.3} / \textbf{52.0}         &   \hspace{\widviss}977.0 (897.2$^{*}$)    \\
\bottomrule
\end{tabular}
}
\caption{Semantic segmentation performance comparison on ADE20K. ``IN1K'' represents ImageNet pre-train. ``CLIP+Semantic FPN'' denotes finetuning with the pre-trained image encoder of CLIP as the backbone. Following DenseCLIP~\cite{rao2022denseclip}, we report the mIoU (\%) of both single-scale and multi-scale testing on both sides of ``/''. We also calculate the FLOPs with 1024 $\times$ 1024 input size via the {\tt fvcore} library. $^{*}$ represents the FLOPs calculated without the text encoder.}
\label{tbl:ade20k}
\vspace{-.1in}
\end{table}

\subsection{Implementation Details}
We implement all our models using the MMSegmentation toolbox~\cite{mmseg2020}. We use Semantic FPN~\cite{kirillov2019panoptic} framework to evaluate our ICPC. Concretely, we use the pre-trained image encoder of CLIP as the backbone, which consists of ResNet-50~\cite{he2016deep} ResNet-101~\cite{he2016deep} and Vision Transformer (ViT-B)~\cite{dosovitskiy2020imagevit}, and use Semantic FPN as the decoder. 
As for prompt learning, we use the pre-trained text encoder of CLIP and use a context size of 9 (8 learnable context vectors and 1 instance-conditioned vector). We use the same transformer decoder architecture as DenseCLIP.
Following the practice of DenseCLIP, we fix the text encoder and finetune the image encoder using a 0.1 learning rate during training.
Unless otherwise specified, we train all the models for 80K iterations using AdamW~\cite{loshchilov2017adamw} with an input size of 512$\times$512.

\subsection{Main Results}

\def\widvis{0.5cm}
\begin{table}[!t]
\centering
\footnotesize 
\resizebox{1.0\linewidth}{!}{
\begin{tabular}{clcll}
\toprule
\multirow{2}{*}{Backbone}              & \multirow{2}{*}{Model}   & \multirow{2}{*}{Pre-train}      & \multicolumn{2}{c}{mIoU (\%)}     \\ \cmidrule{4-5}
                                       &                          &                                 & COCO-Stuff10k              & ADE20K-Full          \\
\midrule
\multirow{4}{*}{ResNet-50}  &   Semantic FPN~\cite{kirillov2019panoptic}  & IN1K                & \hspace{\widvis}31.08      & \hspace{\widvis}9.01 \\
                            &   CLIP+Semantic FPN                         & CLIP                    & \hspace{\widvis}34.19      & \hspace{\widvis}11.64  \\
                            &   DenseCLIP~\cite{rao2022denseclip}         & CLIP                    & \hspace{\widvis}37.79      & \hspace{\widvis}13.27  \\
                            &   ICPC (Ours)                               & CLIP                    & \hspace{\widvis}\textbf{38.84} & \hspace{\widvis}\textbf{14.68}    \\
\midrule
\multirow{4}{*}{ResNet-101} &   Semantic FPN~\cite{kirillov2019panoptic}  & IN1K                & \hspace{\widvis}33.36      & \hspace{\widvis}10.13 \\
                            &   CLIP+Semantic FPN                         & CLIP                    & \hspace{\widvis}37.31      & \hspace{\widvis}13.36  \\
                            &   DenseCLIP~\cite{rao2022denseclip}         & CLIP                    & \hspace{\widvis}40.83      & \hspace{\widvis}14.37 \\
                            &   ICPC (Ours)                               & CLIP                    & \hspace{\widvis}\textbf{41.78} & \hspace{\widvis}\textbf{16.05}    \\ 
\midrule
\multirow{4}{*}{ViT-B}      &   Semantic FPN~\cite{kirillov2019panoptic}  & IN1K                & \hspace{\widvis}43.10      & \hspace{\widvis}16.32 \\
                            &   CLIP+Semantic FPN                         & CLIP                    & \hspace{\widvis}44.46      & \hspace{\widvis}18.09  \\
                            &   DenseCLIP~\cite{rao2022denseclip}         & CLIP                    & \hspace{\widvis}44.45      & \hspace{\widvis}18.37  \\
                            &   ICPC (Ours)                               & CLIP                    & \hspace{\widvis}\textbf{45.38} & \hspace{\widvis}\textbf{19.70}    \\ 
\bottomrule
\end{tabular}
}
\caption{Semantic segmentation performance comparison on COCO-Stuff10k and ADE20K-Full.}
\label{tbl:cocostuff_ade20kfull}
\vspace{-.2in}
\end{table}

\subsubsection{ADE20K}
As shown in Table~\ref{tbl:ade20k}, we compare the semantic segmentation results on ADE20K with three different backbones. In addition to mIoU in single-scale and multi-scale testing, we also report the FLOPs measured by {\tt fvcore} library. The results demonstrate that our ICPC significantly outperforms the comparison methods based on the same backbone. To be specific, ICPC improves the performance by 1.7\%, 2.0\%, and 0.7\% based on ResNet-50, ResNet-101, and ViT-B respectively compared to DenseCLIP. As for FLOPs, since the instance-conditioned prompting introduces text encoder and the align-guided contrastive learning does not introduce any computation during inference, our multi-scale alignment design just introduces 3\%$\sim$5\% FLOPs.

\subsubsection{COCO-Stuff10k}
Table~\ref{tbl:cocostuff_ade20kfull} reports the experimental results on COCO-Stuff10k dataset based on three different backbones. We reimplement the results of the comparison methods based on their released code. We can observe that although DenseCLIP outperforms the vanilla finetuning (CLIP+Semantic FPN) by a large margin, our ICPC can further improve the performance by 1.05\% and 0.95\% mIoU based on ResNet-50 and ResNet-101 respectively. 
It is worth noting that when using ViT-B as backbone, DenseCLIP~\cite{rao2022denseclip} only attains on par result compared to the vanilla finetuning, while our ICPC can further improve the performance from 44.46\% to 45.38\% mIoU.

\subsubsection{ADE20K-Full}
As shown in Table~\ref{tbl:cocostuff_ade20kfull}, we compare ICPC on the more challenging ADE20K-Full dataset, which consists of 847 classes. 
%
Compared to the state-of-the-art method~\cite{rao2022denseclip}, our ICPC also brings consistent improvements, improving the performance by 1.41\%, 1.68\%, and 1.33\% mIoU based on the three backbones, respectively.

\begin{table}[!t]
\centering
\resizebox{1.0\linewidth}{!}{
\begin{tabular}{l|ccc}
\toprule
Prompting Method                            &  Learnable context    & Instance-Conditioned  & mIoU (\%) \\ 
\midrule
{\tt a photo of a [CLASS]}                  &    \XSolidBrush       & \XSolidBrush          & 42.90      \\
CoOp~\cite{zhou2021learning}                &    \Checkmark         & \XSolidBrush          & 43.40      \\
ICPC (Ours)                                 &    \XSolidBrush       & \Checkmark            & 43.44      \\
ICPC (Ours)                                 &    \Checkmark         & \Checkmark            & \textbf{43.88}      \\
\bottomrule
\end{tabular}
}
\caption{Analysis of the effectiveness of instance-conditioned prompting method.
}
\label{tbl:ablation_prompt}
\vspace{-.1in}
\end{table}

\begin{table}[!t]
\centering
\resizebox{1.0\linewidth}{!}{
\begin{tabular}{l|cc}
\toprule
Prompting Method                                      &   Prompting                                   & mIoU (\%) \\ 
\midrule
CoCoOp~\cite{zhou2022cocoop}                &   $[V_1+I, V_2+I, ..., V_N+I, \text{CLS}]$    & 43.68      \\
ICPC (Ours)                                 &   $[V_1, V_2, ..., V_N, I, \text{CLS}]$       & \textbf{43.88}      \\
\bottomrule
\end{tabular}
}
\caption{Prompting comparison: ICPC vs. CoCoOp~\cite{zhou2022cocoop}.}
\label{tbl:ablation_vscocoop}
\vspace{-.2in}
\end{table}

\subsection{Ablation Studies}

\subsubsection{Analysis of Instance-Conditioned Prompting}
This section analyzes the effect of the proposed instance-conditioned prompting. In Table~\ref{tbl:ablation_prompt}, we compare with the fixed and learnable prompting methods. It is worth noting that when only using the instance-conditioned vector ($I$ in Figure~\ref{fig:arch}), we attain \textit{on par} result compared with CoOp~\cite{zhou2021learning} (43.44 vs. 43.40). We achieve the best performance by combining the learnable context and instance-conditioned vector to perform prompting.
Moreover, we find that CoCoOp~\cite{zhou2022cocoop} has used the instance information by adding it to each vector of the learnable context for classification tasks. As reported in Table~\ref{tbl:ablation_vscocoop}, we also reimplement CoCoOp in our framework and find that decoupling the learnable context ($[V_1, V_2, ..., V_N]$) that is shared among all images of the dataset and the instance-conditioned vector ($I$) that encode the instance-specific information can lead to better performance for semantic segmentation tasks.

\subsubsection{Effect of Sampling Strategy}

\begin{table}[!t]
\centering
\resizebox{1.0\linewidth}{!}{
\begin{tabular}{l|cc}
\toprule
Method                                      & Sampling strategy     & mIoU (\%) \\ 
\midrule
\multirow{2}{*}{Align-guided Contrastive Learning}                  &    Random                      & 43.94           \\
                                                                    &    Easy-to-Hard                & \textbf{44.22}            \\
\bottomrule
\end{tabular}
}
\caption{Ablation experiments on the sampling strategy of align-guided contrastive learning.
}
\label{tbl:ablation_sampling}
\end{table}

In this section, we analyze the effect of different sampling strategies of the proposed align-guided contrastive learning method. As reported in Table~\ref{tbl:ablation_sampling}, compared with random sampling, our easy-to-hard sampling strategy can achieve better performance.

\begin{table}[!t]
\centering
\resizebox{1.0\linewidth}{!}{
\begin{tabular}{l|cc}
\toprule
Method                                              & Sampling number               & mIoU (\%) \\ 
\midrule
\multirow{4}{*}{Align-guided Contrastive Learning}  &    2                          & 44.05     \\
                                                    &    5                          & \textbf{44.22}     \\
                                                    &    10                         & 43.81          \\
                                                    &    50                         & 43.61            \\
\bottomrule
\end{tabular}
}
\caption{Ablation experiments on the sampling number of easy-to-hard sampling strategy.
}
\label{tbl:ablation_samplingnumber}
\end{table}

\begin{table}[!t]
\centering
\resizebox{0.5\linewidth}{!}{
\begin{tabular}{l|cc}
\toprule
Method                                      & $\gamma$                      & mIoU (\%) \\ 
\midrule
\multirow{3}{*}{ICPC (Ours)}                &    0.1                        & 44.40          \\
                                            &    0.5                        & \textbf{45.21}            \\
                                            &    1.0                        & 44.96     \\
\bottomrule
\end{tabular}
}
\caption{Ablation experiments on $\gamma$.}
\label{tbl:ablation_gamma}
\vspace{-.2in}
\end{table}

\subsubsection{Effect of Sampling Number}
As reported in Table~\ref{tbl:ablation_samplingnumber}, we analyze the effect of the sampling number by varying the value from 2 to 50.
The results show that we achieve a sweet spot at $|\mathbf{P}|=5$, \ie, sampling 5 positive points per class in a mini-batch can provide better performance.
Besides, we also find that sampling more points may introduce more noise thereby degrading the performance.

\subsubsection{Effect of $\gamma$}
As shown in Table~\ref{tbl:ablation_gamma}, we analysis the effect of $\gamma$ in the overall objective of Equation~(\ref{eq:overallloss}). Our ICPC achieves the best performance by setting $\gamma=0.5$.

\subsubsection{Convergence Analysis}

\begin{figure}[h]
    \centering
    \includegraphics[width=0.94\linewidth]{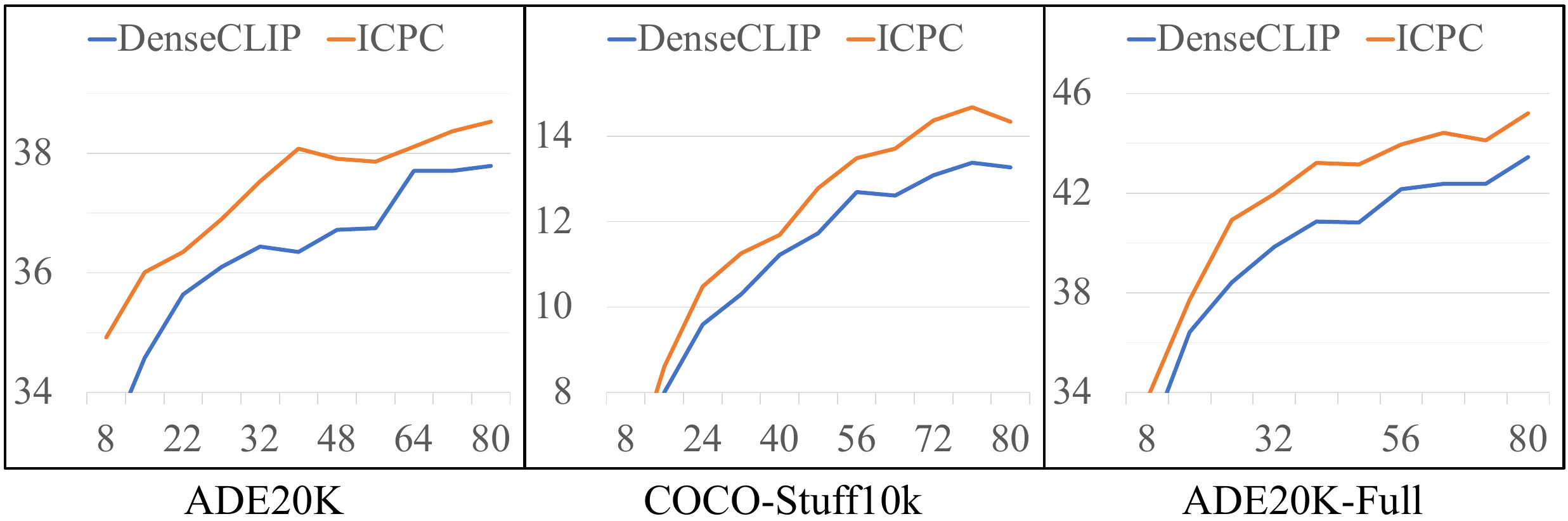}
    \vspace{-.1in}
    \caption{Convergence comparison on three datasets based on ResNet-50. The Y-axis and X-axis of each figure are mIoU (\%) and iterations (K) respectively.}
    \label{fig:ablation_converge}
    \vspace{-.05in}
\end{figure}

As shown in Figure~\ref{fig:ablation_converge}, we report the convergence curve \wrt~training process on three datasets. We can conclude that our ICPC can consistently improve the performance compared to the previous literature~\cite{rao2022denseclip}.

\subsubsection{Factor-by-factor Analysis}

\begin{table}[h]
\centering
\resizebox{0.96\linewidth}{!}{
\begin{tabular}{l|cccccc}
\toprule
\multirow{2}{*}{Method}             & \multirow{2}{*}{I.C.}     & \multirow{2}{*}{C.L.} & \multirow{2}{*}{M.S.}  & \multirow{2}{*}{mIoU (\%)} & \multicolumn{2}{c}{GFLOPs} \\ \cmidrule{6-7} 
                                    &                           &                       &                        &                            & Text encoder & Other modules \\
\midrule
DenseCLIP~\cite{rao2022denseclip}   &                           &                    &                 & 43.50          &  74.3   & 269.7  \\ 
\midrule
\multirow{9}{*}{ICPC (Ours)}        &                           &                    &                 & 43.40          &  74.3   & 269.7  \\ 
                                    &    \Checkmark             &                    &                 & 43.88          &  79.8   & 269.9  \\
                                    &                           &   \Checkmark       &                 & 44.22          &  74.3   & 269.7  \\
                                    &                           &                    &   \Checkmark    & 43.70          &  74.3   & 284.1  \\
                                    &    \Checkmark             &                    &   \Checkmark    & 44.00          &  79.8   & 284.2  \\
                                    &                           &   \Checkmark       &   \Checkmark    & 44.73          &  74.3   & 284.1   \\
                                    &    \Checkmark             &   \Checkmark       &                 & 44.90          &  79.8   & 269.9   \\
                                    &    \Checkmark             &   \Checkmark       &   \Checkmark    & \textbf{45.21} &  79.8   & 284.2   \\

\bottomrule
\end{tabular}
}
\caption{Ablation experiments on the three proposed modules of ICPC. `I.C.', `C.L.', and `M.S.' represent instance-conditioned prompting, align-guided contrastive learning and multi-scale alignment, respectively.}
\label{tbl:ablation_ade20k}
\end{table}

As shown in Table~\ref{tbl:ablation_ade20k}, we conduct a factor-by-factor experiment on our proposed instance-conditioned prompting, align-guided contrastive learning, and multi-scale vision-text alignment.
We also provide the FLOPs in detail.
Each component has a positive effect, and all components are combined to obtain the best performance.





\subsection{Zero-shot Cross-domain Dataset Performance}
To demonstrate the robustness and generalization of our method, we compare on four zero-shot datasets. Concretely, we first train all the methods on COCO-Stuff10k, then we directly evaluate the performance of shared classes on other cross-domain datasets. COCO-Stuff10k shares 17 classes, 45 classes, 12 classes, and 3 classes with Pascal VOC~\cite{everingham2015pascalvoc}, Pascal Context~\cite{mottaghi2014pascalcontext}, Cityscapes~\cite{cordts2016cityscapes}, and iSAID~\cite{waqas2019isaid}, respectively.
As shown in Table~\ref{tbl:zeroshot}, our ICPC achieves better performance.

\begin{table}[!t]
\centering
\footnotesize 
\resizebox{1.0\linewidth}{!}{
\begin{tabular}{l|cccc|c}
\toprule
\multirow{2}{*}{Method}                         & \multicolumn{5}{c}{Evaluation on cross-domain data (mIoU)}      \\ \cmidrule{2-6}
                                                & VOC        & Pascal Context    & Cityscapes       & iSAID          & Mean \\
\midrule
Semantic FPN~\cite{kirillov2019panoptic}        & 66.29             & 40.12             & 32.64            & 6.37           & 36.36  \\
CLIP+Semantic FPN                               & 70.38             & 43.65             & 36.43            & 15.49          & 41.49  \\
DenseCLIP~\cite{rao2022denseclip}               & 76.85             & 46.33             & 37.07            & 17.93          & 44.55  \\
ICPC (Ours)                                     & \textbf{78.25}    & \textbf{47.03}    & \textbf{37.73}   & \textbf{18.74} & \textbf{45.44}  \\
\bottomrule
\end{tabular}
}
\caption{Quantitative comparison on zero-shot cross-domain datasets.}
\label{tbl:zeroshot}
\vspace{-.1in}
\end{table}

\subsection{Can ImageNet pre-trained backbone align with CLIP pre-trained Text Encoder?}

\begin{table}[!t]
\centering
\footnotesize 
\resizebox{0.9\linewidth}{!}{
\begin{tabular}{lll}
\toprule
Decoder                       & Method                 & \hspace{0.5cm}mIoU (\%) \\
\midrule
\multirow{6}{*}{Semantic FPN}     & ResNet-50              & \hspace{0.5cm}38.60      \\
                                                              & ResNet-50+DenseCLIP~\cite{rao2022denseclip}    & \hspace{0.5cm}41.00      \\
                                                              & ResNet-50+ICPC (Ours)  & \hspace{0.5cm}\textbf{43.43 (+2.43)}  \\ \cmidrule{2-3}
                                                              & ResNet-101             & \hspace{0.5cm}40.40      \\
                                                              & ResNet-101+DenseCLIP~\cite{rao2022denseclip}   & \hspace{0.5cm}43.00      \\
                                                              & ResNet-101+ICPC (Ours) & \hspace{0.5cm}\textbf{46.44 (+3.44)}        \\
\bottomrule
\end{tabular}
}
\caption{Applying ICPC to ImageNet pre-trained backbones on ADE20K dataset.}
\label{tbl:anybackbone}
\vspace{-.2in}
\end{table}

As reported in Table~\ref{tbl:anybackbone}, we replace the pre-trained image encoder of CLIP with ImageNet pre-trained models to verify the multimodal alignment capability of ICPC. The results demonstrate that ICPC can bring significant performance improvements against DenseCLIP. It is worth noting that our ICPC 
finetuned on ImageNet pre-trained models
achieves \textit{on par} or even better performance compared to DenseCLIP finetuned using CLIP's image encoder.

\begin{figure}[!t]
    \centering
    \includegraphics[width=0.95\linewidth]{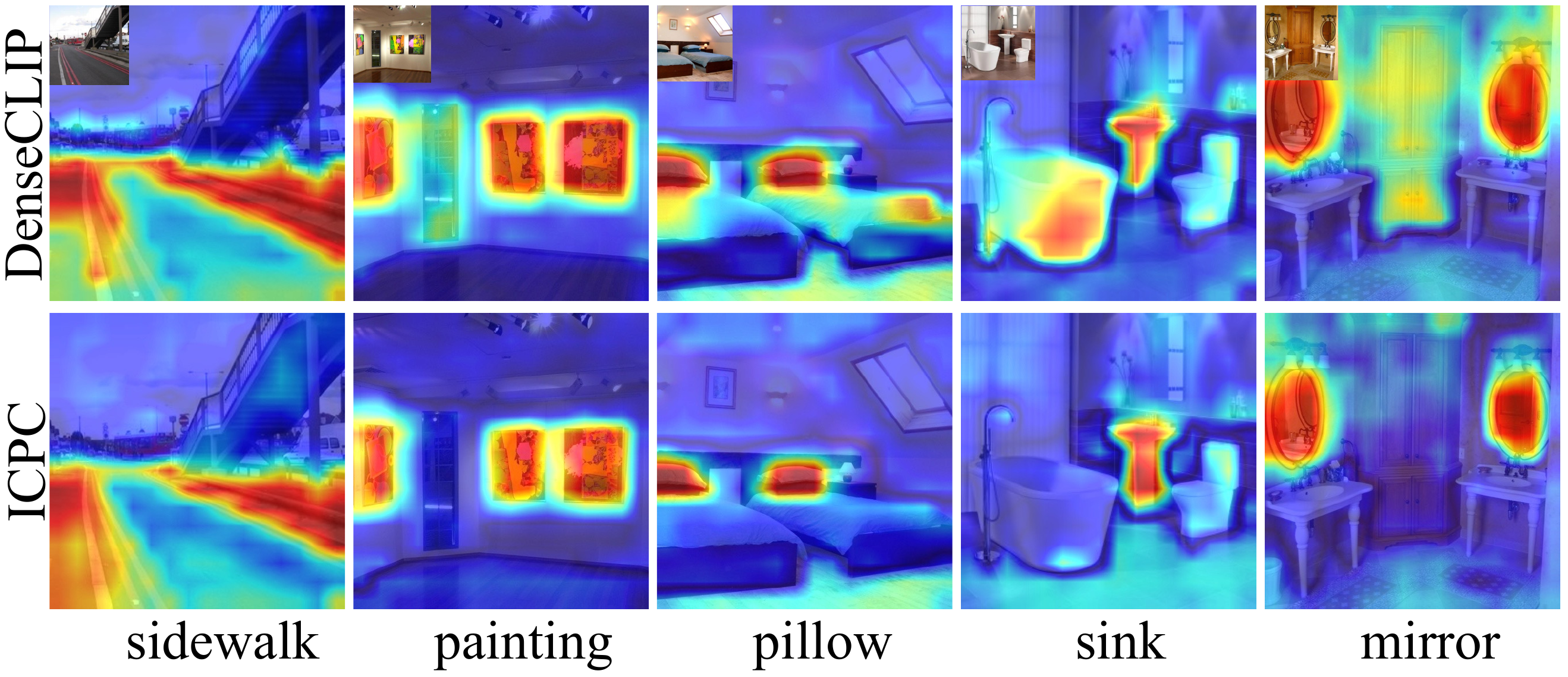}
    \vspace{-.1in}
    \caption{Visualization of class-specific regions of the aligned multimodal embeddings.}
    \label{fig:vis_scoremap}
    \vspace{-.1in}
\end{figure}

\begin{table}[!t]
\centering
\resizebox{0.98\linewidth}{!}{
\begin{tabular}{lcll}
\toprule
\multirow{2}{*}{Method}             & \multirow{2}{*}{Backbone}   & \multicolumn{2}{c}{mIoU (\%)} \\ \cmidrule{3-4}
                                    &                             & \hspace{0.5cm}Decoder           & \hspace{0.5cm}Multimodal embedding $\mathcal{A}$ \\
\midrule
DenseCLIP~\cite{rao2022denseclip}   & \multirow{2}{*}{ResNet-50}  & \hspace{0.5cm}43.50             & \hspace{0.5cm}39.06  \\
ICPC (Ours)                         &                             & \hspace{0.5cm}\textbf{45.21 (+1.71)}    & \hspace{0.5cm}\textbf{41.44 (+2.38)}  \\                  
\bottomrule
\end{tabular}
}
\caption{Performance comparison by using the outputs of the decoder and the aligned multimodal embeddings $\mathcal{A}$ as the segmentation results, respectively.}
\label{tbl:eval_scoremap}
\vspace{-.15in}
\end{table}

\subsection{Visualization}
We visualize the class-specific regions of the aligned multimodal embeddings learned on ADE20K~\cite{zhou2019semanticade20k} in Figure~\ref{fig:vis_scoremap}, the backbone is ResNet-50 and the weights of DenseCLIP come from the official released model. We can see that ICPC generates clear and high attention scores for the correct objects. 
Besides, we directly use the aligned multimodal embeddings $\mathcal{A}$ as segmentation results to evaluate the alignment quality. ICPC achieves superior performance improvements (+2.38\% mIoU), indicating that our ICPC can effectively improve the quality of vision-text alignment when transferring CLIP to downstream semantic segmentation tasks. 
We also provide visualization samples of the segmentation results in our \texttt{supplementary} materials.


\section{Discussion and Limitation}
Although our ICPC framework has achieved consistent performance improvements when transferring CLIP to downstream segmentation tasks, the computational cost would be the major limitation. 
The reason is that,
compared with static prompt designs of DenseCLIP that the text embeddings can be computed in advance, our instance-conditioned prompting requires a feed-forward through the text encoder to obtain the image-specific text embeddings, which can not be computed before inference.
%
%
We believe that this issue can be further improved by designing a more efficient prompting strategy.

\section{Conclusion}
In this paper, we present an instance-conditioned prompting with contrastive learning framework, named ICPC, to more effectively transfer the knowledge of pre-trained vision-language models (CLIP) to downstream semantic segmentation tasks. 
We propose instance-conditioned prompting, align-guided contrastive learning, and multi-scale vision-text alignment to better align the vision-text modalities, which is extremely helpful for downstream semantic segmentation tasks.
Extensive experiments demonstrate the superior performance of our method.

\bibliography{aaai23}


\end{document}